\documentclass[lettersize,journal]{IEEEtran}
\usepackage{amsmath,amsfonts}
\usepackage{algorithmic}
\usepackage{algorithm}
\usepackage{array}
\usepackage[caption=false,font=normalsize,labelfont=sf,textfont=sf]{subfig}
\usepackage{textcomp}
\usepackage{stfloats}
\usepackage{url}
\usepackage[letterpaper, left=0.75in, right=0.75in, top=1in, bottom=0.75in]{geometry}
\usepackage{verbatim}
\usepackage{graphicx}
\usepackage{cite}
\usepackage{adjustbox}
\usepackage{orcidlink}
\usepackage{color}
\hypersetup{
    colorlinks=false,
    pdfborder={0 0 0}
}

\begin{document}
\title{A Novel Aerial-Aquatic Locomotion Robot with Variable Stiffness Propulsion Module}
\author{Junzhe Hu\textsuperscript{1,2}\orcidlink{0009-0001-4902-9424}, Pengyu Chen\textsuperscript{*,1}\orcidlink{0009-0007-7137-5714}, Tianxiang Feng\textsuperscript{*,1}\orcidlink{0009-0002-5298-9837}, Yuxuan Wen\textsuperscript{1}\orcidlink{0000-0002-6017-2117}, Ke Wu\textsuperscript{3}, Janet Dong\textsuperscript{1}\orcidlink{0000-0002-5607-1727}~\IEEEmembership{}

\thanks{\textsuperscript{1}Junzhe Hu, Pengyu Chen, Tianxiang Feng, Yuxuan Wen, and Janet Dong are with the UC Center for Robotics Research, College of Engineering and Applied Science, University of Cincinnati, Cincinnati, OH, USA (e-mail: hujz@mail.uc.edu, chenp7@mail.uc.edu, fengtn@mail.uc.edu, wenyu@mail.uc.edu, dongjg@ucmail.uc.edu).}
\thanks{\textsuperscript{2}Junzhe Hu is with the Robotics Institute, School of Computer Science, Carnegie Mellon University, Pittsburgh, PA, USA (e-mail: junzhehu@andrew.cmu.edu).}
\thanks{\textsuperscript{3}Ke Wu is with Robotics department, Mohamed bin Zayed University of Artificial Intelligence, Masdar City, Abu Dhabi, United Arab Emirates (e-mail:Ke.Wu@mbzuai.ac.ae).}
\thanks{*: Authors with equal contribution..}
}


\maketitle

\begin{abstract}
In recent years, the development of robots capable of operating in both aerial and aquatic environments has gained significant attention. This study presents the design and fabrication of a novel aerial-aquatic locomotion robot (AALR). Inspired by the diving beetle, the AALR incorporates a biomimetic propulsion mechanism with power and recovery strokes. The variable stiffness propulsion module (VSPM) uses low melting point alloy (LMPA) and variable stiffness joints (VSJ) to achieve efficient aquatic locomotion while reduce harm to marine life. The AALR's innovative design integrates the VSPM into the arms of a traditional quadrotor, allowing for effective aerial-aquatic locomotion. The VSPM adjusts joint stiffness through temperature control, meeting locomotion requirements in both aerial and aquatic modes. A dynamic model for the VSPM was developed, with optimized dimensional parameters to increase propulsion force. Experiments focused on aquatic mode analysis and demonstrated the AALR's swimming capability, achieving a maximum swimming speed of 77 mm/s underwater. The results confirm the AALR's effective performance in water environment, highlighting its potential for versatile, eco-friendly operations. 
\end{abstract}


\section{Introduction}
\IEEEPARstart{C}{onventional} UAVs and AUVs are widely utilized in rescue operations, environmental monitoring, and protection. To diversify the tasks performed by robots, research groups have shifted their focus to the study of trans-media robots~\cite{ref1,ref2,ref3}. Existing research indicates that there are currently three primary methods employed to achieve trans-media locomotion between air and water~\cite{ref4}. One method involves using a propeller on the robot's head to provide kinetic energy and sufficient lift during flight, with deformable fixed wings reducing drag in water and aiding in attaining yaw, roll, and pitch angles~\cite{ref5}. This method has demonstrated significant advantages, such as improved maneuverability and the ability to adapt to different media seamlessly. Researchers at ETH Zurich~\cite{ref6} have shown that this approach allows for precise control and stability during transitions between air and water, which is crucial for environmental monitoring.

Another approach involves converting chemical and thermal energy into kinetic energy to help the robot leap out of the water. This method has been effective in providing rapid transitions and burst movements necessary for specific applications, such as escaping underwater obstacles or quickly deploying to aerial positions. Research conducted by Imperial College London~\cite{ref7}\cite{ref8} has highlighted the efficiency and reliability of this method in diverse environmental conditions.

Most research groups are exploring air-water locomotion using multi-rotor UAVs due to their simple water entry and exit control and stable structure. However, traditional UAV propellers pose risks to marine life and are prone to entanglement with disturbances such as seaweed, resulting in motion loss of the trans-media robots. Additionally, due to the difference in fluid density between aerial and aquatic environments, conventional propellers typically achieve only 5\% propulsive efficiency underwater~\cite{ref1}.

To address these challenges, a novel AALR capable of efficient movement in both aerial and aquatic environments has been developed. Certain diving beetles can achieve high-speed locomotion in both air and water. They utilize the passive deformation of hairs on their swimming legs to generate effective propulsion force by exploiting the drag difference, achieving underwater propulsion efficiency of up to 84\%~\cite{ref9}. Inspired by these beetles, the concept of passive deformation is adopted in the design of the paddles. To increase the range of the blade's swimming motion, the blades are rigidly connected to a variable stiffness continuum structure driven by cables. This approach not only provides effective propulsion but also reduces the number of drive units and simplifies control~\cite{ref10}.

Achieving cross-medium locomotion without traditional propellers necessitates additional mechanisms for aquatic movement, which presents significant challenges to the UAV's flight efficiency and payload capacity~\cite{ref11}\cite{ref12}. LMPA offers a promising solution in this research direction~\cite{ref13}. To minimize the impact of additional mechanism weight on AALR flight, the rigid frame of a traditional quadcopter has been transformed into a VSPM using VSJ which consists of LMPA and silicon rubber, a cable-driven continuum structure, and a heating system. In aerial mode, the LMPA remains solid, providing sufficient rigidity to maintain flight stability. In aquatic mode, the LMPA is heated, turning the VSJ into flexible joints. The VSPM, through the cable-driven continuum structure combined with passive joint paddles, achieves wide-range rowing motion for effective aquatic propulsion without significantly increasing the overall mass. The design essentials are shown in Figure. 1.

This innovative design enhances the propulsion ability and adaptability of the AALR in underwater environments through biomimetic propulsion modes inspired by the diving beetle's swimming legs, while also reducing harm to marine life~\cite{ref14}. This study makes three main contributions: (1) A novel AALR with VSPM using VSJ which consists of LMPA and silicon rubber has been designed and fabricated, enabling effective aerial-aquatic locomotion without traditional propellers and adding aquatic locomotion ability with low additional weight. (2) Inspired by diving beetles, the AALR adopts a biomimetic rowing propulsion mode with power and recovery strokes, reducing harm to marine life, and has efficient propulsion ability. The design integrates passive joint paddles and a cable-driven continuum structure for enhanced propulsion and simplified control. (3) Through six experiments, the AALR achieved a swimming speed of 77 mm/s in aquatic mode. The heating system effectively melts LMPA in VSJ for the aquatic mode with reliable performance. To the best of our knowledge, this work is the first to replace the frame of a traditional quadcopter with a VSPM to achieve aerial-aquatic locomotion.

The rest of this paper is organized as follows: Section II delves into the design of the AALR. Section III details the prototype, its components, and the dynamic modeling. In Section IV, the experimental results for both aquatic and aerial modes are presented and discussed. Finally, Sections V and VI offer conclusions and propose directions for future work.

\begin{figure}
\centering
\setlength{\fboxrule}{1pt} 
\setlength{\fboxsep}{0pt}  
\fbox{\includegraphics[width=3in]{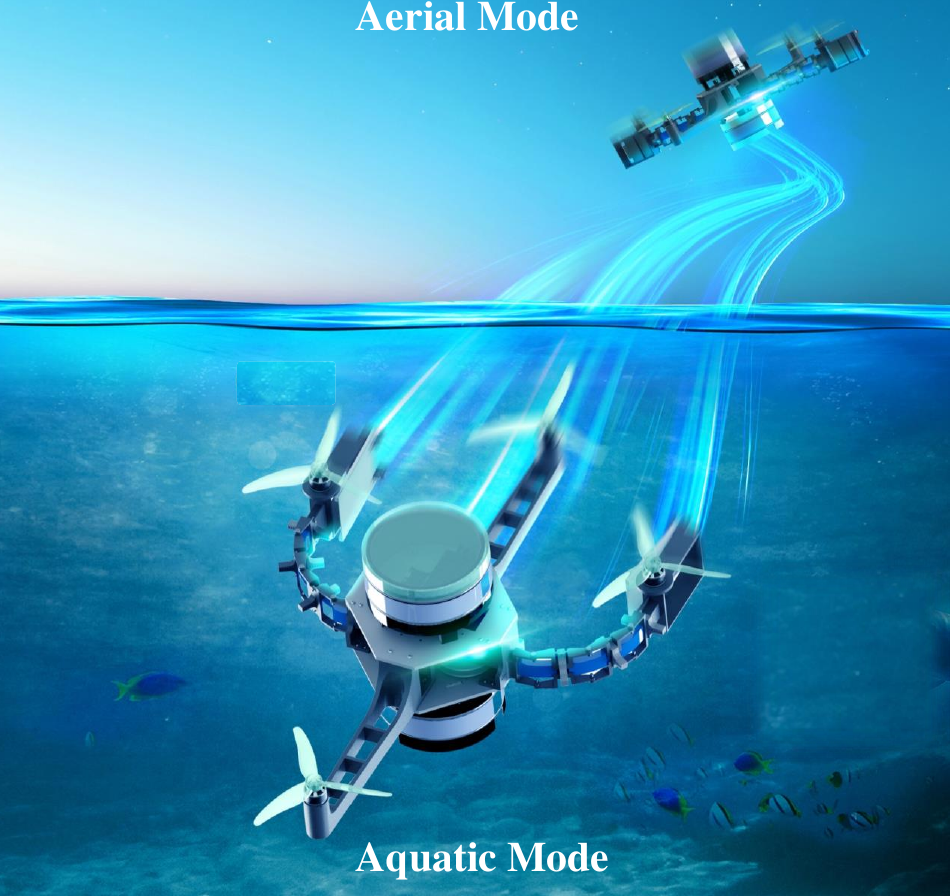}}
\caption{{Design essentials of the AALR prototype.} }
\label{fig8}
\end{figure}

\section{Aerial Aquatic Locomotion Robot Design}

To ensure the AALR's capability to perform rowing motion underwater while protecting the electronic components from water exposure, multiple waterproofing strategies were adopted, which was inspired by aquatic UAV design\cite{ref15}. These include the use of waterproof sealants such as epoxy resin and silicone sealant, surface coatings like acrylic paint, and waterproof gaskets. The waterproof tank~\cite{ref16}, designed to house the electronics controller, has an interior volume of 4050 mm$^3$ and consists of six main parts:
\begin{enumerate}
\item Gasket: 3D printed using TPU.
\item Cap1 and Cap2: 3D printed using PLA.
\item Outer cover and Hull: 3D printed using PLA.
\item O-Ring: Sourced from Amazon.
\end{enumerate}

Sealing efficacy is achieved through four primary contact points:
\begin{enumerate}
\item Contact between the outer cover and the hull.
\item Contact between the hull and the main cap.
\item Contact between the hull and the gasket.
\item Contact between the gasket and the main cap.
\end{enumerate}

To validate the waterproof capability of the AALR, a waterproofing experiment was conducted. A smaller-scale tank with the same sealing design as the actual tank, costing less than \$30, was used. The tank was submerged to a depth of 500 mm in a water tank. The experiment demonstrated an effective waterproof duration of 70 minutes. The comprehensive and detailed structure of the waterproof tank is illustrated in Figure. 2ab.
 

\begin{figure}[H]
\centering
\includegraphics[width=3in]{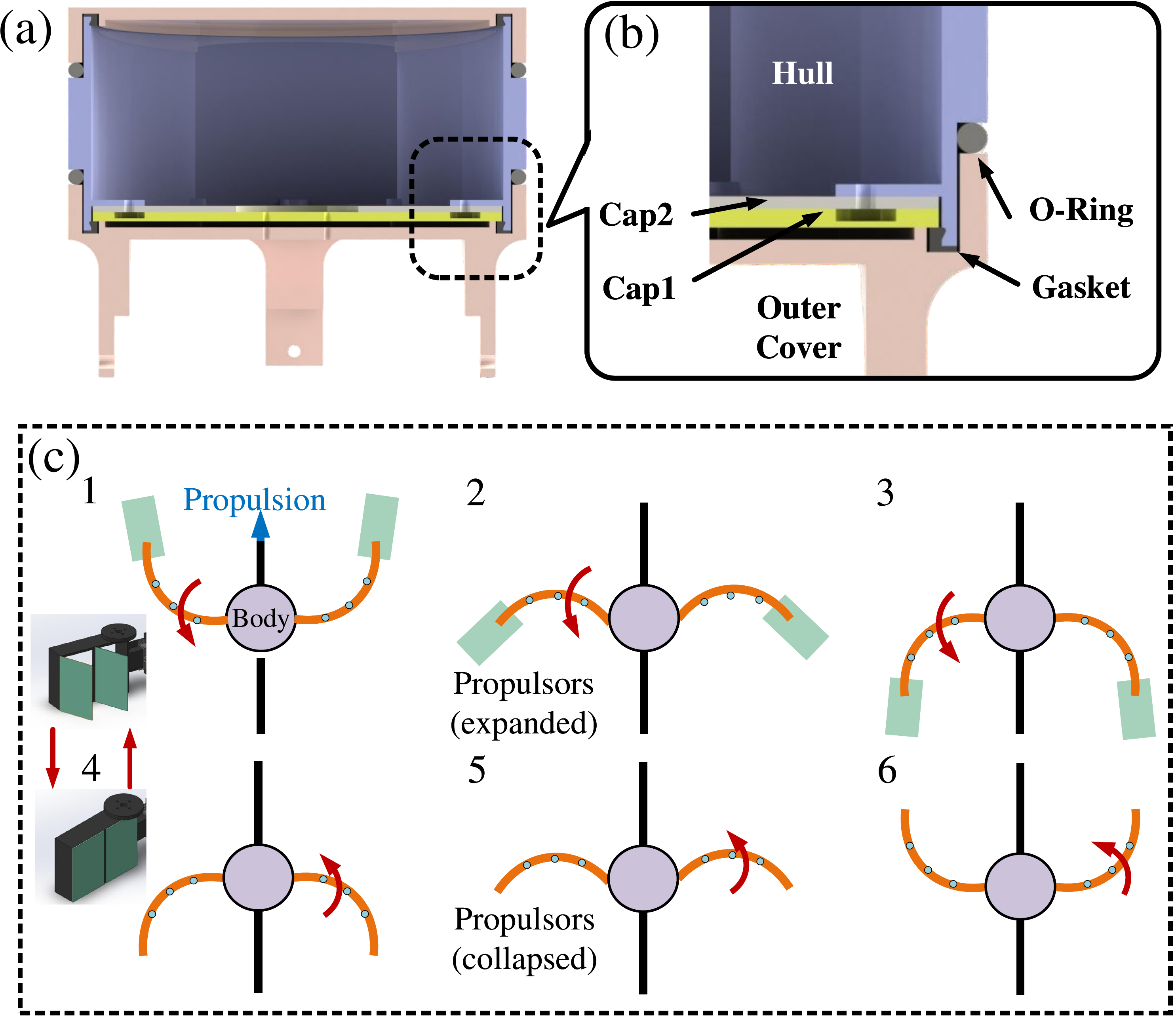}
\caption{{Experimental setup} (\textbf{a}) Cross-section of the entire waterproof tank. (\textbf{b}) Enlarged image of six parts. (\textbf{c}) Schematic of propulsion period, 1-3: power stroke; 4-6: recovery stroke.}
\label{fig8}
\end{figure}

Inspired by the diving beetle, which achieves underwater propulsion efficiency of up to 84\%, the propulsion mechanism mimics its swimming leg's motion and passive deformation hair. The propulsion cycle consists of two primary strokes: the power stroke (Figure. 2c 1-3) and the recovery stroke (Figure. 2c 4-6). During the power stroke, the passive joint paddle interacts with the fluid, presenting a large force projection area. In the recovery stroke, the paddle interacts with the fluid on the opposite side, passively opening and significantly reducing the projection area. This dynamic allows the paddle to achieve a positive net propulsion force in each cycle~\cite{ref17}. The changes in the force projection area of the passive joint paddle during these strokes are depicted in Figure. 2c. By integrating these design elements, the AALR efficiently operates underwater while maintaining the integrity of its electronic components, thereby enhancing its overall performance and reliability in diverse environments.


\section{Prototype}

\subsection{Variable Stiffness Joint Design}

The functionality of the VSJ is achieved using LMPA, and we used only the gallium component in this work. When the temperature is below the melting point, LMPA remains solid and exhibits good stiffness. When the temperature exceeds the melting point, LMPA becomes liquid and flexible. The components and materials of the VSJ are shown in Figure. 3 and Table 1. Based on the design, the AALR operates in two main modes:
\begin{enumerate}
    \item In aerial mode, the UAV arm is rigid to ensure flight stability.
    \item In aquatic mode, the UAV arm becomes flexible, allowing for effective rowing motion.
\end{enumerate}

Inspired by research on swimming robots and soft actuators\cite{ref10}\cite{ref13}, a VSJ was designed with the following capabilities:
\begin{enumerate}
    \item In aerial environments, the VSJ remains rigid to maintain flight stability.
    \item In aquatic environments, the VSJ becomes flexible to facilitate swimming.
\end{enumerate}

\begin{figure}[H]
\centering
\includegraphics[width=3.2in]{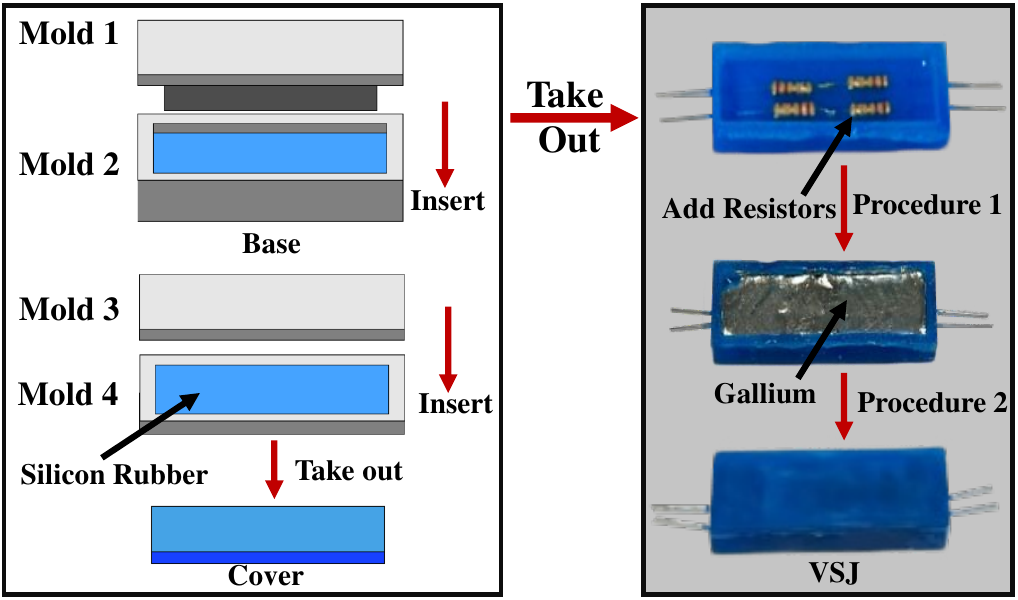}
\caption{Component and material of VSJ.}
\label{fig8}
\end{figure}
\vspace{-0.3cm}
\begin{table}[htbp]
\caption{Components and materials of VSJ}
\centering
\begin{tabular}{>{\centering\arraybackslash}m{1.3cm}>{\centering\arraybackslash}m{1.1cm}>{\centering\arraybackslash}m{1.1cm}>{\centering\arraybackslash}m{1.7cm}>{\centering\arraybackslash}m{1.1cm}}
\hline
\textbf{Component} & \textbf{Material} & \textbf{Density (g/cm$^3$)} & \textbf{Young’s Modulus (Pa)} & \textbf{Poisson's Ratio} \\
\hline
Cover & Silicon Rubber & 1.2 & $3 \times 10^6$ & 0.45 \\
\hline
Metal & Gallium & 5.91 & 9.8 & 0.31 \\
\hline
Base & Silicon Rubber & 1.2 & $3 \times 10^6$ & 0.45 \\
\hline
\end{tabular}
\label{table:component_material}
\end{table}

A VSJ is made of 4 parts, including silicon rubber cover, gallium metal layer, silicon rubber layer with resistor inside, and silicon rubber base, as shown in Fig. 3. It is fabricated layer by layer:
\begin{enumerate}
    \item Arrange four resistors in a 2x2 configuration at the bottom of base. Pour silicon rubber into the base to submerge the whole resistor and liquid gallium metal to cover silicon rubber layer.
    \item Glue the cover and base together.
\end{enumerate}

\begin{figure}[H]
\centering
\includegraphics[width=3.2in]{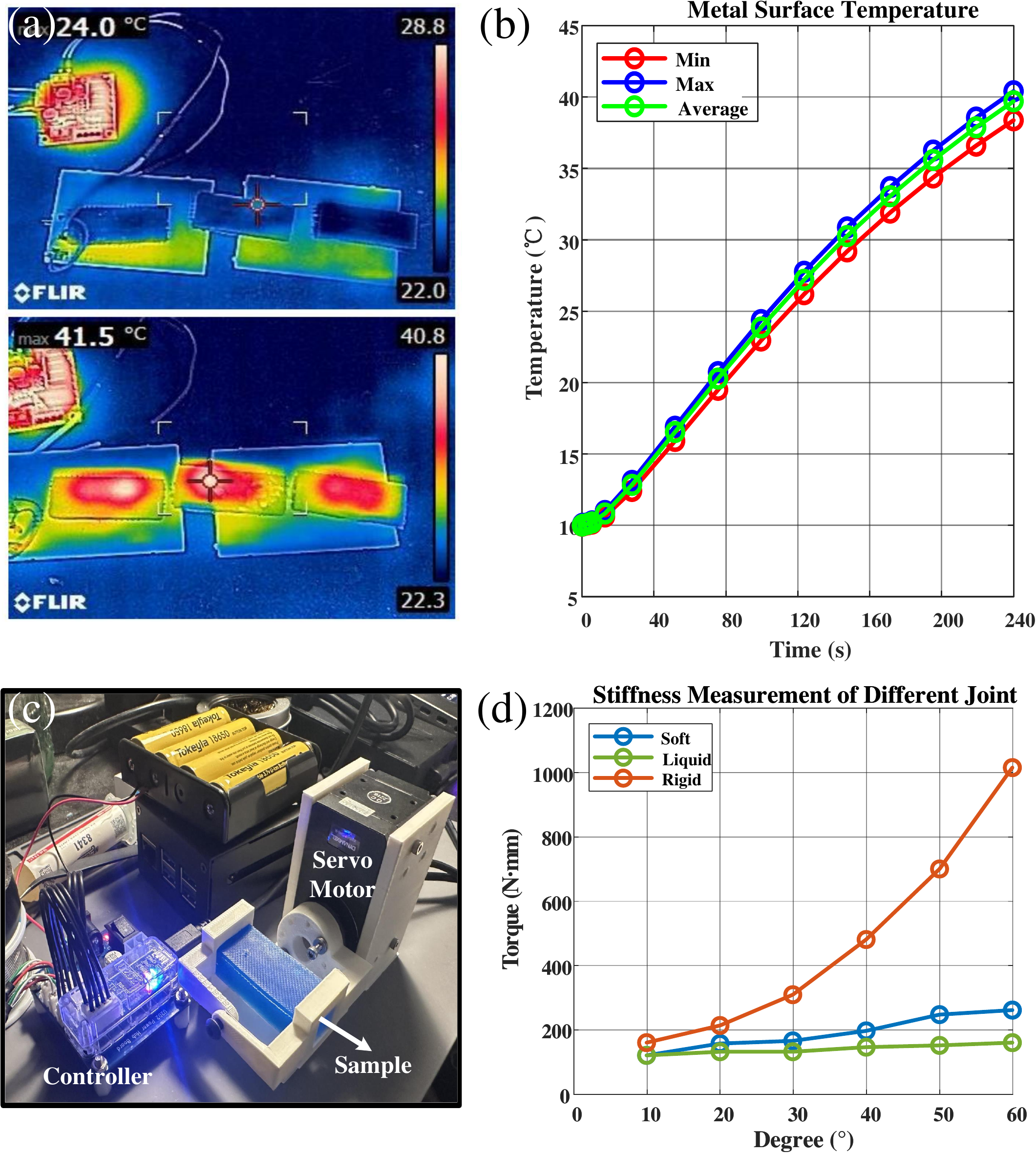}
\caption{Stiffness measurement experiments of joints. (\textbf{a}) Thermal imaging results. (\textbf{b}) Metal surface temperature plot. (\textbf{c}) Stiffness measurement experimental setup. (\textbf{d}) Stiffness measurement of soft, rigid, and liquid joints.}
\label{fig8}
\end{figure}

To validate the heating system of the AALR, a thermal imaging camera was used to visualize the temperature changes of the VSJs, as shown in Figure. 4a. Three VSJs were fabricated and connected in series, mimicking the joint connection in the AALR. The temperature changes of the VSJs are depicted in Figure. 4b, indicating it takes about 200 seconds for the VSJs to heat from 10\textdegree C to 40\textdegree C. The melting temperature of gallium metal was 30\textdegree C, confirming that the metal inside the joints had completely melted.

To characterize the stiffness changes of the VSJ, a servomotor and an experimental platform were utilized, as shown in Figure. 4c. One end of the joint was fixed to a base, while the other end was connected to a rotating shaft. As the servomotor drove the rotating shaft, the joint bent under the applied force. The rotation angle (\textdegree) was input into the servomotor, and the current value was recorded from the software.

The stiffness of a spring is defined as $k = \frac{\Delta F}{\Delta X}$, where \( \Delta F \) represents the force difference and \( \Delta X \) represents the displacement difference. For the VSJ, the stiffness is defined as $K_J = \frac{\Delta \tau}{\Delta \theta}$, where \( \Delta \tau \) represents the torque change measured in experiments, and \( \Delta \theta \) represents the angle difference input to the servomotor. By knowing the relationship between current and torque, the torque on the joint was calculated based on the measured current value. Figure. 4d shows that the joint with solid gallium metal had the highest stiffness, while the joint with liquid gallium metal had the lowest stiffness. This experiment confirmed that the rigid joint had significantly greater stiffness compared to the other two joints, aligning with the original design intention.

\subsection{Variable Stiffness Propulsion Module}
In the aquatic mode of AALR, under the same control parameters, the cable-driven continuum structure has the advantages of a large range of motion, high compliance, and reduced risk of harm to marine life upon contact. This helps to limit the VSJ’s position during the rigid-flexible transition transformation. In this research, the cable-driven continuum is selected to accomplish the propulsive action. The main structure of VSPM is shown in Figure. 5a below. It has three VSJs (in blue) to supply enough rotation angle in propulsion motion. A waterproof servomotor is fixed on one side with a rigidly connected reel. Two cables are rigidly fixed at each end of a reel, and each cable passes through predetermined holes in a continuum structure consisting of three links. The cables are rigidly attached to the final link, which is equipped with a passive joint paddle at the end. This cable-driven continuum structure enhances the paddles' oscillation range and motion profile, effectively improving propulsion with a limited number of control parameters (amplitude in degree). The dimensional parameters of the VSPM were optimized by dynamic modeling, as shown in Figure. 5b, with the parameters listed in Table 2. Figure. 5c illustrates the schematic of the rigid and flexible states of the VSPM~\cite{ref17}.

\begin{figure*}[htbp!]
\centering
\includegraphics[width=7in]{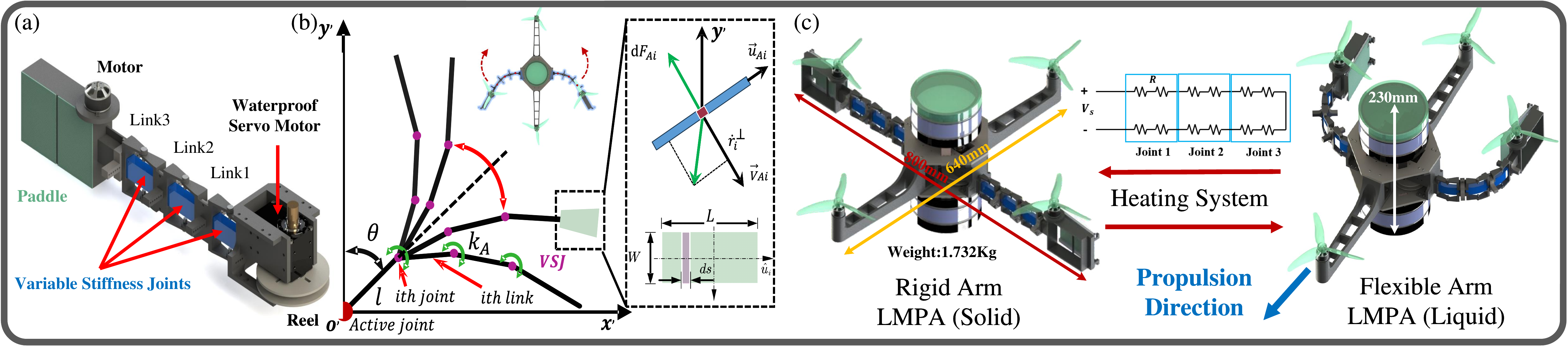}
\caption{ (\textbf{a}) Structure of VSPM. (\textbf{b}) Schematic diagram of VSPM dynamic modeling. (\textbf{c}) Change joint stiffness utilizing the heating system and LMPA.}
\label{fig8}
\end{figure*}

\begin{table}
\renewcommand{\arraystretch}{1.5}
\caption{Key parameters in dynamic modeling for VSPM.}
\label{tab:parameters}
\centering
\begin{tabular}{>{\centering\arraybackslash}p{1.5cm}>{\centering\arraybackslash}p{5cm}>{\centering\arraybackslash}p{1.2cm}}
\hline
\hline
\textbf{Symbol} & \textbf{Meaning} & \textbf{Value} \\
\hline
\hline
$N$ & Number of the links & 3 \\
\hline
$\theta$ & Motor's rotation angle (°)& - \\
\hline
$l_1 = T + h_0$ & Distance between each joint (mm) & 110 \\
\hline
$h_0$ & Distance between two links before rotating (mm) & 80 \\
\hline
$u_Ai$ & Unit vector parallel to the $i^{th}$ link & - \\
\hline
$v_Ai$ & Unit vector perpendicular to the $i^{th}$ link & - \\
\hline
$\beta$ & Distance between the specific point (mm) & - \\
\hline
$r_{Ai}(\beta)$ & Position of any point at the $i^{th}$ link (mm) & - \\
\hline
$\dot{r}_{Ai}(\beta)$ & Velocity (mm/s) & - \\
\hline
$\dot{r}^{\perp}_{Ai}(\beta)$ & Perpendicular projection of the velocity (mm/s) & - \\
\hline
$F_{Ai}(\beta)$ & Hydrodynamic force (N) & - \\
\hline
$C_n$ & Hydrodynamic coefficient & - \\
\hline
$L$ & Length of paddle (mm) & 90 \\
\hline
$W$ & Width of paddle (mm) & 60 \\
\hline
\end{tabular}
\end{table}

${u}_{Ai}$ and ${v}_{Ai}$ are unit vectors parallel and perpendicular to the $i^{th}$ link:
\begin{equation}
{\textbf{u}}_{\textbf{Ai}} = 
\begin{bmatrix}
\cos \left( \frac{iy}{N} \right) \\
\\
\sin \left( \frac{iy}{N} \right)\\
\end{bmatrix}
\end{equation}

\begin{equation}
{\textbf{v}}_{\textbf{Ai}} = 
\begin{bmatrix}
-\sin \left( \frac{iy}{N} \right) \\
\\
\cos \left( \frac{iy}{N} \right)
\end{bmatrix}
\end{equation}

In the $x'o'y'$ coordinate, the position of any point at the $i^{th}$ link of the cable-driven body, ${r}_{Ai}(\beta)$, can be represented as:

\begin{equation}
{\textbf{r}}_{\textbf{Ai}}(\beta) = 
\begin{bmatrix}
\displaystyle\sum_{j=0}^{i-1} \left( l_1 \cos \left( \frac{jy}{N} \right) \right) + \beta \cos \left( \frac{iy}{N} \right) - l_1 \\
\displaystyle\sum_{j=0}^{i-1} \left( l_1 \sin \left( \frac{jy}{N} \right) \right) + \beta \sin \left( \frac{iy}{N} \right)
\end{bmatrix}
\end{equation}
where $\beta$ is the distance between the specific point and the $i$-th joint, $i=1,2,\ldots,N$. The velocity, $\dot{{r}}_{Ai}(\beta)$, of any point at the $i^{th}$ link of the cable-driven active body can be obtained:

\begin{equation}
\dot{{\textbf{r}}}_{\textbf{Ai}}(\beta) = 
\begin{bmatrix}
-\frac{l_1 \dot{y}}{N} \displaystyle\sum_{j=0}^{i-1} \left( \sin \left( \frac{jy}{N} \right) \right) - \beta \frac{i \dot{y}}{N} \sin \left( \frac{iy}{N} \right) \\
\frac{l_1 \dot{y}}{N} \displaystyle\sum_{j=0}^{i-1} \left( \cos \left( \frac{jy}{N} \right) \right) + \beta \frac{i \dot{y}}{N} \cos \left( \frac{iy}{N} \right)
\end{bmatrix}
\end{equation}

Thus, the perpendicular projection of the velocity with respect to the $i^{th}$ link is:

\begin{equation}
\dot{{\textbf{r}}}_{\textbf{Ai}}(\beta) = (\dot{{r}}_{Ai}(\beta) \cdot {v}_{Ai}) {v}_{Ai}
\end{equation}

The hydrodynamic force is shown as:

\begin{equation}
\textbf{F}_{\textbf{Ai}}(\beta) = \int_{0}^{L} -C_h W \rho \left|\dot{{r}}_{Ai}(\beta)\right|\left|\dot{{r}}_{Ai}(\beta)\right| \, ds
\end{equation}

To simplify the complexity of the dynamics model of the VSPM, we assume that the delay of the passive paddle is fixed at 0.1 seconds when it switches its position between the power stroke and the recovery stroke. The dynamic system of VSPM was constructed in Simulink, and the propulsion force ($F_y$) simulation results were counted (control parameters: 0.2Hz, $\theta$ = 60°, offset: 0°). According to the simulation results, the dimensional data were adjusted for a total of 30 iterations, of which three representative ones are shown in Figure. 6.

\begin{figure}[H]
\centering
\includegraphics[width=3.4in]{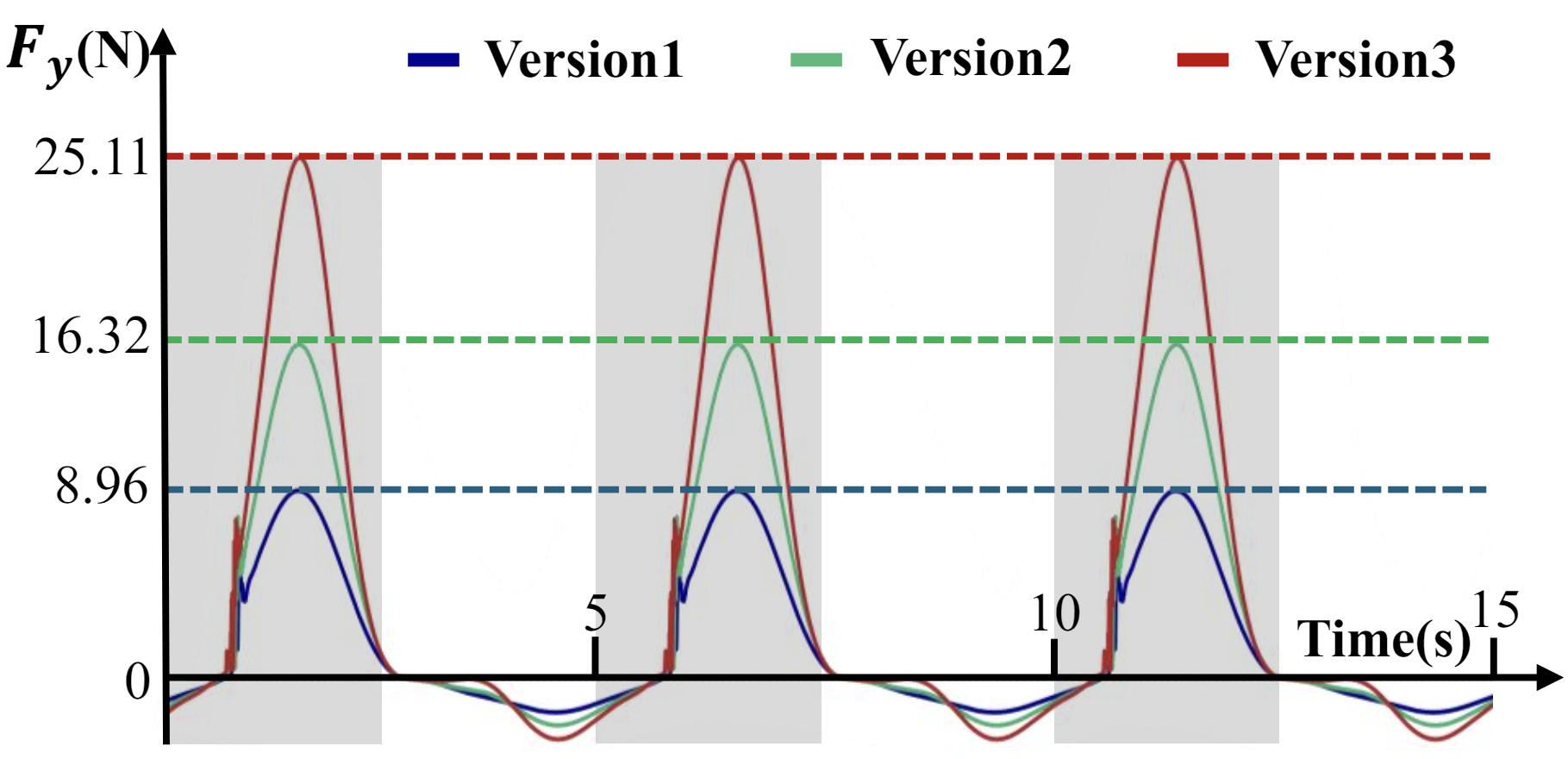}
\caption{Results of propulsion force in each simulation for three versions of dimensions design (control parameters: 0.2Hz, $\theta$ = 120°, offset: 0°).}
\label{fig8}
\end{figure}

In aerial mode, a high-stiffness arm is required to ensure the stability of the Multirotor UAV during flight. The VSPM needs to stop heating the Gallium in the VSJs to allow itself to cool down and transform from a liquid to a solid state, thus increasing the stiffness of the joints. During the conversion process, the waterproof motor needs to return to its initial position and be limited by cables to ensure that each link of the Rigid Arm state is parallel to the other.

\subsection{Control System}

The control system of the AALR is composed of three main parts: the heating system, the aquatic mode, and the aerial mode. The components of the control system are shown in Figure. 7 and Table 3.

\begin{figure}
\centering
\includegraphics[width=3.2in]{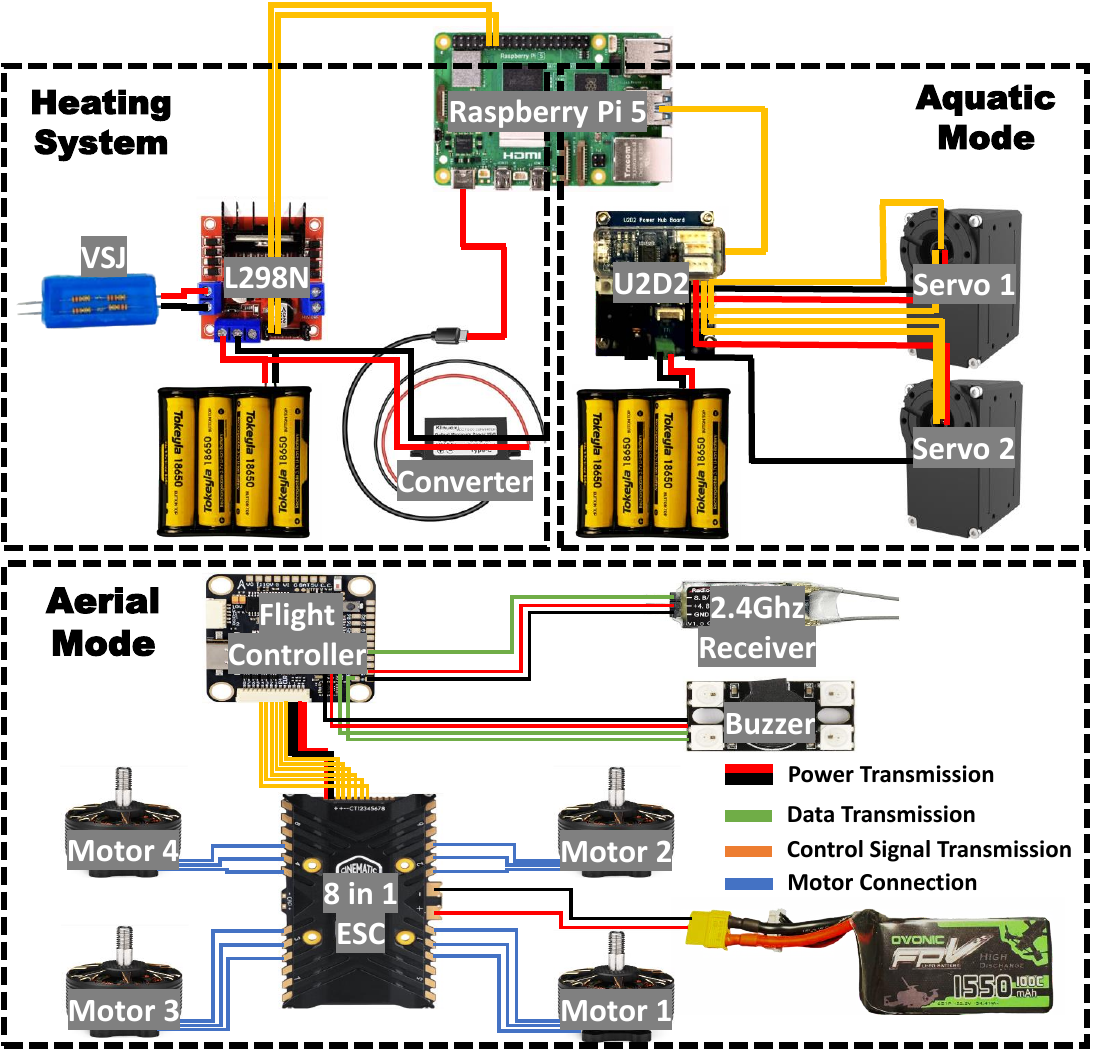}
\caption{Components of the control system.}
\label{fig8}
\end{figure}

\begin{table}
\centering
\caption{Components of the control system.}
\begin{tabular}{>{\centering\arraybackslash}m{2cm}>{\centering\arraybackslash}m{6cm}}
\hline
\hline
\textbf{Item} & \textbf{Model} \\
\hline
\hline

\textbf{Heating System} & \\
\hline
Controller & Raspberry Pi 5 \\
\hline
Battery & Tokeyla 18650 Rechargeable Battery 4000mAh \\
\hline
Converter & Klnuoxj DC 12V/24V to 5V USB C Step Down Converter \\
\hline
Control Panel & L298N \\
\hline
\textbf{Aquatic Mode} & \\
\hline
Controller & Raspberry Pi 5 \\
\hline
Battery & Tokeyla 18650 Rechargeable Battery 4000mAh \\
\hline
Converter & Klnuoxj DC 12V/24V to 5V USB C Converter \\
\hline
Control Panel & ROBOTIS U2D2 \\
\hline
Servo & ROBOTIS DYNAMIXEL XW540-T140-R \\
\hline
\textbf{Aerial Mode} & \\
\hline
Motor & T-Motor CINE66 2812-KV1155 \\
\hline
Propeller & T-Motor C7.5*4.6 \\
\hline
Electrical Speed Controller & T-Motor CINE55A 8S 8IN1 32BIT \\
\hline
Flight Controller & T-Motor F7 Pro \\
\hline
Battery & OVONIC 6S LiPo Battery 100C 1550mAh \\
\hline
Transmitter & Radiolink AT9S Pro \\
\hline
Receiver & Radiolink R12DSM 2.4Ghz \\
\hline
Buzzer & iFlight 5V Buzzer \\
\hline
\end{tabular}
\label{table:components_heating_system}
\end{table}

The heating system is used to melt the LMPA inside the VSJs. The principle of the heating system follows Joule's law:

\begin{equation}
Q = I^{2}Rt
\end{equation}
where the current \( I \) and the resistance \( R \) are known, and the only parameter to control is the time \( t \). Therefore, a Raspberry Pi 5 is used as the controller to manage the input power supply, turning it on and off as needed. The heating operation is completed before the swimming motion of the AALR, and the system will continue working during the motion to keep VSJs soft.
\begin{figure*}[hbb!]
\centering
\includegraphics[width=6.5in]{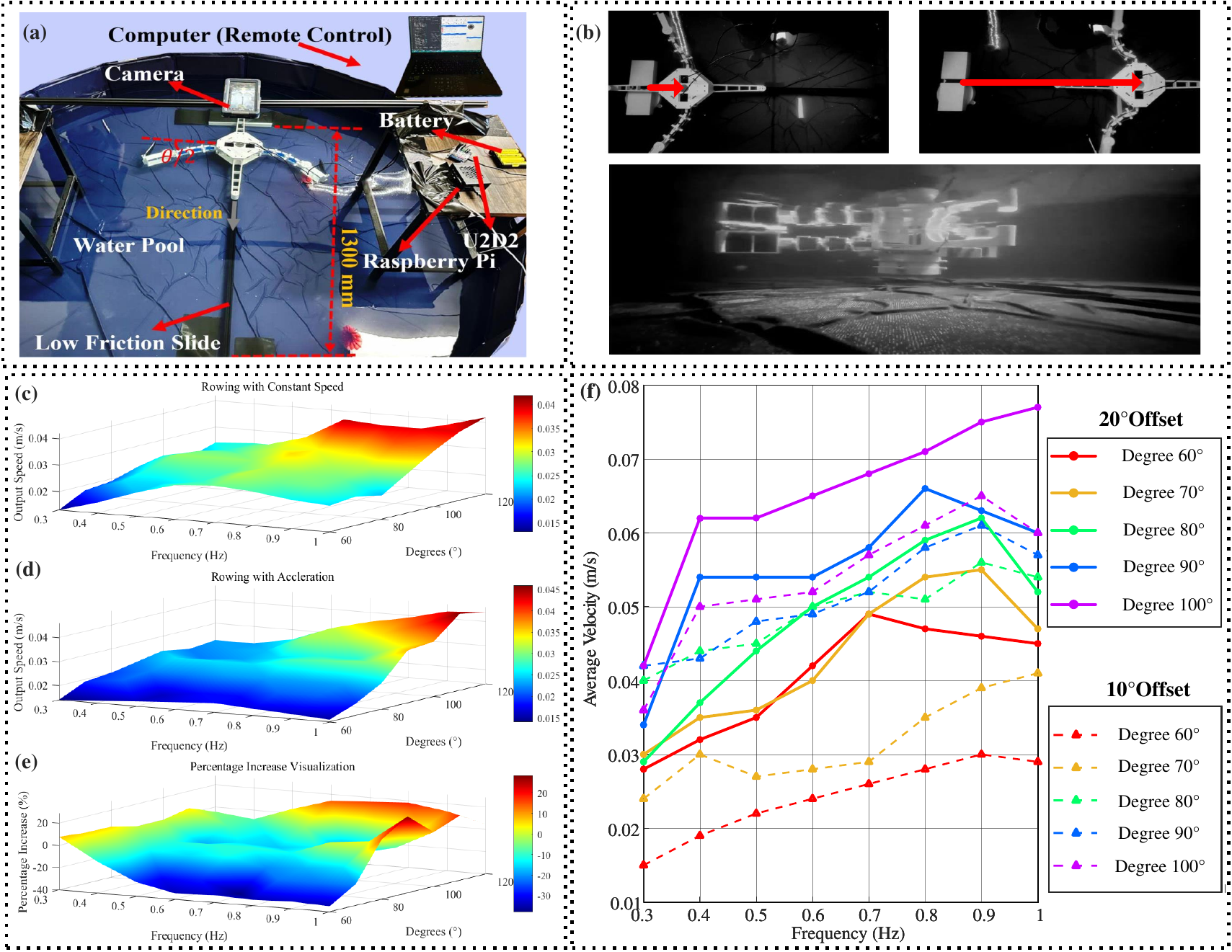}
\caption{Experimental results. (\textbf{a}) Experiment platform for propulsion speed test. (\textbf{b}) Swimming speed distribution (rowing at constant speed). (\textbf{c}) Swimming speed distribution (rowing at constant speed). (\textbf{d}) Experimental procedure demonstration. \textbf{Top: }Comparison graph of AALR swimming speed at baseline and optimal control parameters. \textbf{Bottom:} Underwater side view of the experimental process. (\textbf{e}) The speed difference between constant speed and acceleration. (\textbf{f}) Swimming speed distribution (offset = 10° and 20°).}
\label{fig8}
\end{figure*}
The aquatic mode controls the swinging motion of the VSPM to achieve underwater movement. The Raspberry Pi 5 controls the input parameters of the servos to generate the swinging motion of the VSPM. The control parameters will be discussed in detail in Sec. IV.

\begin{figure}[H]
\centering
\includegraphics[width=2.2in]{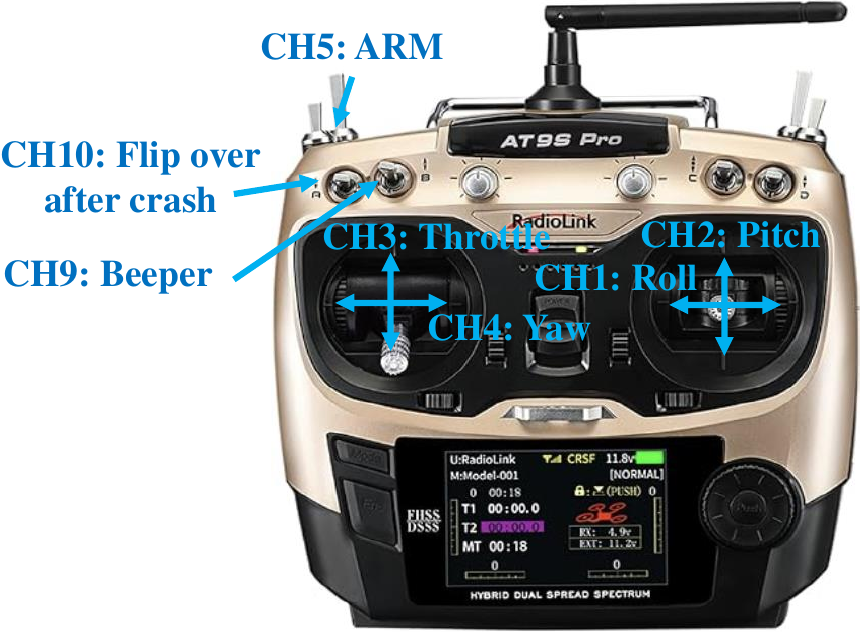}
\caption{Channels of the transmitter.}
\label{fig8}
\end{figure}

The aerial mode facilitates the flying of the AALR~\cite{ref11}. A transmitter is used to control the AALR, which is powered independently by AA batteries. The channels of the transmitter are shown in Figure. 8.

There are four basic parameters controlled through the four main channels: the throttle, the yaw angle, the pitch angle, and the roll angle. The throttle and the yaw angle are controlled by the left rocker, while the pitch angle and the roll angle are controlled by the right rocker. Additionally, there are other parameters to control. Channel 5, the rocker on the left rear, controls ARM, which is the lock of the AALR. Channels 9 and 10, the rockers on the left front, control the beeper and the function of flipping over after a crash.

\section{Results and Discussion}

\subsection{Experiment of Aquatic Mode}
The swimming speed of the AALR in aquatic mode was tested, and the effects of control parameters on swimming speed were verified. The control parameters included rowing amplitude ($\theta$, °), frequency ($f$, Hz), acceleration ($\alpha$, °/$s^2$), and asymmetry (offset, °). The experimental platform consisted of a 1500 mm long rail set up in a 2000 mm diameter pool. A low-friction slide was rigidly fixed to the base of the AALR, and a camera was used to record the motion of the AALR and the advancement of the VSPM. A Raspberry Pi and a U2D2 were used to remotely control the AALR's control parameters from a PC. The pool's diameter was more than five times the length of the VSPM, and there was a 30-second interval between each experiment to reduce water perturbation interference. Each experiment for the control parameters was averaged five times to minimize the impact of external environmental factors. The experimental platform is shown in Figure. 9a. A comparison of AALR swimming speed at baseline and optimal control parameters and the underwater side view of the experimental process is shown in Figure. 9b.

Firstly, the swimming speed of the AALR for different rowing amplitudes ($\theta$, °) and frequencies ($f$, Hz) was tested. The control parameters ranged from $\theta \in (60^\circ \sim 120^\circ)$ and $f \in (0.3 \sim 1.0 \ \text{Hz})$. The experimental results are shown in Figure. 9c, with the maximum propulsive velocity of 0.042 m/s achieved when $\theta$ = 120° and $f$ = 1.0 Hz. The results indicated that the swimming velocity of the AALR was positively correlated with rowing amplitude and frequency.

Subsequently, the swimming speed of the AALR with acceleration ($\alpha$ = 100°/s$^2$) was tested. The control parameters ranged from $\theta \in (60^\circ \sim 120^\circ)$ and $f \in (0.3 \sim 1.0 \ \text{Hz})$. The results, shown in Figure. 9d, indicated a maximum propulsive velocity of 0.046 m/s when $\theta$ = 110° and $f$ = 1.0 Hz were used. The application of acceleration generally increased the average swimming speed of the AALR by 5\% to 30\% under the same control parameters, as shown in Figure. 9e. The increase in average swimming speed ranged from -25\% to 30\%, which was related to the delay in the passive deformation of the joint paddles. The acceleration of the VSPM oscillations effectively increased the maximum propulsive speed of the AALR.

Finally, the swimming speed of the AALR with asymmetry (offset of 10° and 20°) was tested. The control parameters ranged from $\theta \in (60^\circ \sim 120^\circ)$ and $f \in (0.3 \sim 1.0 \ \text{Hz})$. The results, shown in Figure. 9f, indicated a maximum propulsive velocity of 0.077 m/s when $\theta$ = 110°, $f$ = 1.0 Hz, and an offset of 20° were used. The application of asymmetry generally increased the average swimming speed of the AALR under the same control parameters. The average swimming speed increased by 10\% to 30\% with the offset applied, with the highest swimming speed of 77 mm/s achieved at a 20° offset, a 592\% improvement compared to the baseline of 13 mm/s. This speed significantly exceeded the success criteria of 60 mm/s.

\subsection{Experiment of Aerial Mode}

Aerial mode testing verified the position control ability and reliability of the VSJs (solid-state) on the VSPM. Due to the battery capacity (1500 mAh), the AALR has an estimated maximum flight time of 15.3 minutes per flight. Given that our work primarily focuses on the variable stiffness mechanism using LMPA and the validation of aquatic locomotion capabilities, the aerial mobility of the UAV, which is a more mature application, has been preliminarily tested. In our future work, we aim to further enhance the flight stability of the UAV and conduct a more detailed quantitative analysis of its flight performance. Particular focus is on control strategies in trans-media processes, and the experiment is shown in Figure 10. 

\begin{figure}[H]
\centering
\includegraphics[width=2.5in]{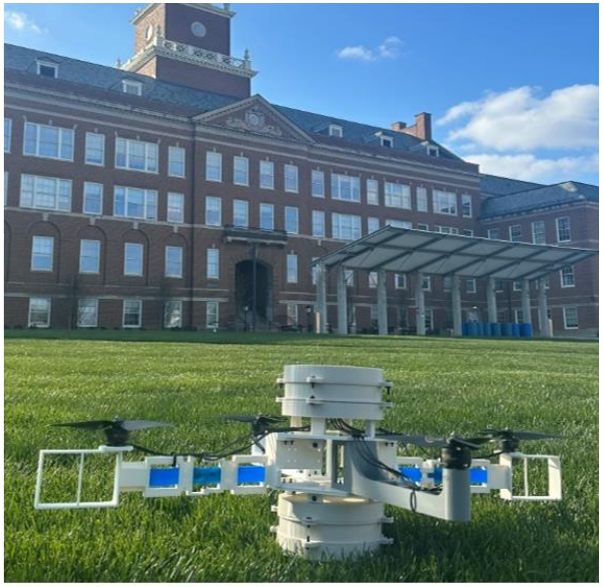}
\caption{Experiment of aerial mode in the campus.}
\label{fig8}
\end{figure}  

\section{Conclusion}
This work introduces a novel concept for an AALR, showcasing its potential for efficient movement in both aerial and aquatic environments. Since the flight control for quadrotor UAVs is relatively mature, this research focuses primarily on the variable stiffness mechanism and its application in aquatic mode. By integrating a VSPM utilizing LMPA-based VSJ, the AALR achieves efficient propulsion and exceptional adaptability in both mediums. Inspired by the diving beetle, this biomimetic propulsion mechanism enables effective propulsion while the compliant propeller minimizes potential harm to marine life. Experimental results demonstrated the robot's capability to achieve a swimming speed of 77 mm/s and a flight duration of 15.3 minutes, validating the functionality and reliability of the proposed design. The use of VSPM and VSJ enhances the robot's ability to seamlessly transition between air and water, maintaining structural rigidity during flight and flexibility during aquatic operations. This unique mechanism reduces the complexity and weight of additional propulsion systems, highlighting its superiority for cross-medium UAV applications. The AALR offers significant potential for applications in rescue missions, environmental monitoring, and marine research, providing a versatile and eco-friendly solution for multi-environment operations. 

\section{Future work}
In the future, our work will focus on developing robust aerial-to-aquatic and aquatic-to-aerial control strategies for the AALR. These strategies aim to reduce energy loss during water entry and exit, ensuring smoother transitions and enhanced stability. Additionally, we will optimize the robot's gravitational distribution to improve flight stability, making it more efficient and reliable in aerial mode.

\vfill

\end{document}